\documentclass{article}

 \usepackage[preprint]{corl_2026} 
\usepackage{graphicx}
\usepackage{amsmath} 
\usepackage{booktabs}
\usepackage{multirow} 
\usepackage{amssymb}
\usepackage{tabularx}
\title{RAPT: Model-Predictive Out-of-Distribution Detection and Failure Diagnosis for Sim-to-Real Humanoid Deployment}

\author{
Humphrey Munn\\
School of Electrical Engineering and Computer Science\\
University of Queensland, QLD 4072, Australia\\
CSIRO Technology, Australia\\
\texttt{[h.munn@uq.edu.au]}\\
\And
Brendan Tidd\\
CSIRO Technology, Australia\\
\And
Peter B{\"o}hm\\
CSIRO Technology, Australia\\
\And
Marcus Gallagher\\
School of Electrical Engineering and Computer Science\\
University of Queensland, QLD 4072, Australia\\
\And
David Howard\\
CSIRO Technology, Australia
}

\begin{document}
\maketitle


\begin{abstract}
Deploying learned control policies is risky because policies that appear robust in simulation can confidently enter out-of-distribution (OOD) states after Sim-to-Real transfer, causing silent failures and potential hardware damage. Existing anomaly detectors often fail to meet the requirements of high-rate control, extremely low false-positive rates, and interpretable failure feedback. We present \textbf{RAPT} (Recurrent Anomaly Probabilistic Trajectory Model), a lightweight, self-supervised $50$\,Hz deployment monitor that learns nominal execution from large-scale simulation and produces calibrated, per-dimension predictive-deviation signals online. RAPT enables OOD detection under strict false-positive constraints while localizing \emph{when and where} real execution departs from nominal behavior. For post-hoc diagnosis, RAPT combines temporal saliency, joint-kinematic summaries, and LLM-based semantic reasoning to classify likely failure causes in a zero-shot setting. In simulation across four Isaac Lab tasks, RAPT improves TPR by $37\%$ over the strongest baseline at $0.5\%$ episode-level FPR; on hardware, it achieves $89\%$ TPR across $78$ trials with fewer false positives than high-frequency-compatible baselines, and reaches $75\%$ semantic failure diagnosis accuracy across $21$ categories on a challenging OOD subset. Project website, code, and videos: \url{https://humphreymunn.github.io/RAPT/}.

\end{abstract}



\section{Introduction}

\begin{figure}[t]
    \centering
    \includegraphics[width=0.85\linewidth]{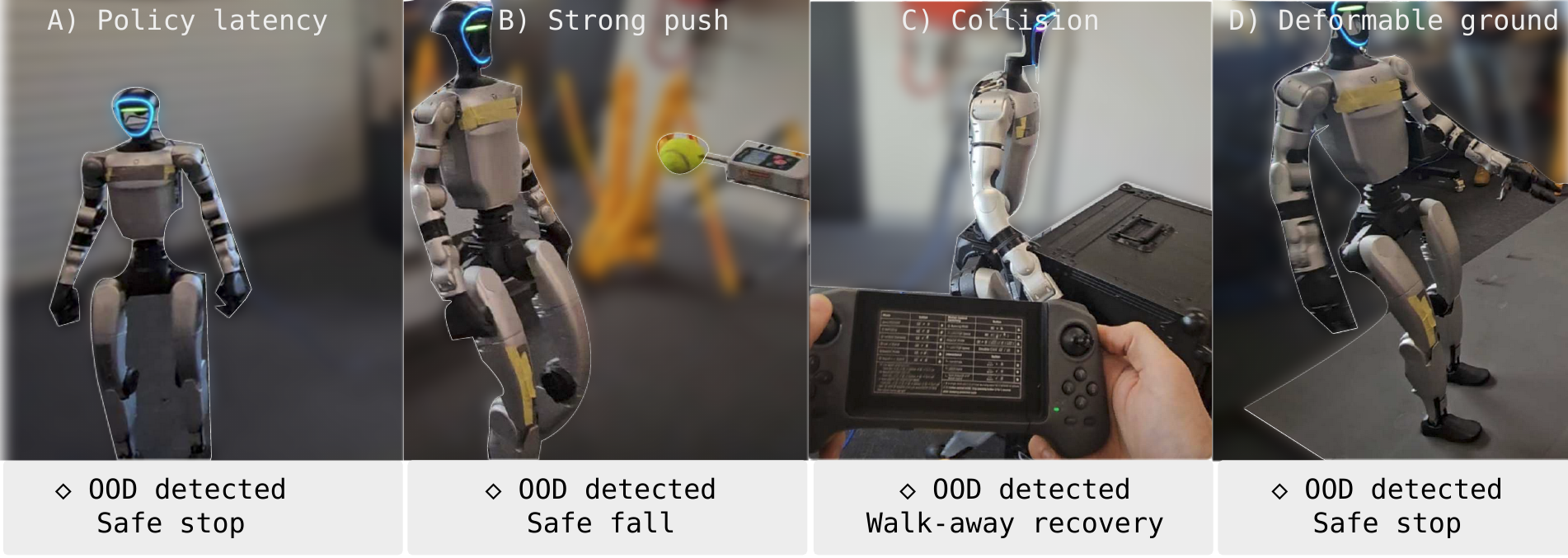}
    \caption{\textbf{RAPT overview.} Real-world out-of-distribution (OOD) scenarios during humanoid deployment. RAPT detects anomalies online and triggers predefined safety responses, including safe stopping, controlled falling, and recovery behaviors.}
    \label{fig:rapt_scenario}
    \vspace{-1em}
\end{figure}

Deploying learned humanoid control policies is risky: policies that appear robust in simulation can confidently enter out-of-distribution (OOD) states after Sim-to-Real transfer, causing silent failures and potential hardware damage~\citep{mohammed2021benchmark,haider2024can,haider2023out}. These failures can arise from unmodeled dynamics, sensor noise, latency, payload changes, terrain variation, or implementation errors, and are particularly concerning for high-dimensional humanoids where unsafe actions can rapidly lead to falls, self-collisions, or hardware faults. Figure~\ref{fig:rapt_scenario} illustrates representative real-world OOD scenarios encountered during humanoid deployment, along with the corresponding safety responses triggered by RAPT.

Existing anomaly detectors do not fully satisfy the requirements of deployment-time humanoid monitoring. Uncertainty and ensemble methods can be computationally expensive at control rate, generative models such as LSTM-VAEs can suffer from likelihood miscalibration and input-complexity bias~\citep{park2018multimodal,nalisnick2018deep}, and generic time-series methods often treat proprioceptive channels as independent signals rather than coupled robot states~\citep{sellam2025mamba,zhong2025patchad}. Most importantly, many approaches provide only a binary stop signal, offering limited insight into \emph{when}, \emph{where}, or \emph{why} execution departed from nominal behavior.

In this work, we introduce \textbf{RAPT} (Recurrent Anomaly Probabilistic Trajectory Model), a lightweight deployment-time monitoring and diagnostic system for humanoid robots. RAPT runs alongside a pre-trained control policy and learns a probabilistic model of nominal spatio-temporal behavior from simulation. During deployment, it evaluates predictive deviation as calibrated, per-dimension signals, enabling online OOD detection under strict false-positive constraints while localizing when and which proprioceptive channels depart from nominal behavior. By optionally calibrating these signals on a small amount of verified nominal real-world data, RAPT can suppress static Sim-to-Real bias while still flagging new real-world deviations such as collisions, payload changes, terrain shifts, sensor faults, or actuator degradation. RAPT complements domain randomization and system identification: rather than replacing methods that reduce the Sim-to-Real gap before deployment, it monitors the residual mismatch and unforeseen OOD events that remain during real execution. This provides an observable deployment-time signal for safety monitoring and debugging without modifying the underlying controller.

Our primary contributions are:
\begin{itemize}
    \item \textbf{Simulation-trained, deployment-calibrated monitoring:} RAPT learns nominal humanoid behavior from large-scale simulation and calibrates thresholds using either simulation or brief verified real-world data, supporting both Sim-to-Real debugging and hardware-time OOD detection.

    \item \textbf{Low-FPR OOD detection for high-DOF humanoids:} RAPT uses probabilistic predictive-deviation modeling with calibrated per-dimension, global, and optional range gates, outperforming high-frequency-compatible baselines under strict episode-level false-positive constraints.
    
    \item \textbf{Localized evidence for failure diagnosis:} RAPT produces per-dimension negative log-likelihood (NLL) signals and temporal saliency through the recurrent model, enabling post-hoc semantic diagnosis without labeled failure data during detector training.
\end{itemize}

We validate RAPT in NVIDIA Isaac Lab and on a physical Unitree G1. In simulation, RAPT improves TPR by $37\%$ over the strongest baseline at $0.5\%$ episode-level FPR; on real hardware, it achieves $89\%$ TPR across $78$ trials while supporting semantic diagnosis of challenging OOD failures. Upon acceptance, we will publicly release the training pipeline, labeled real-world datasets, and C++ deployment code to support reproducibility and future benchmarking.


	

\section{Related Work}
\label{sec-related}

OOD detection in reinforcement learning remains fragmented in terminology and evaluation protocols~\citep{prashant2025guaranteeing,nasvytis2024rethinking}. Unlike supervised OOD detection, robotic deployment anomalies may arise from sensory corruptions, software errors, unmodeled dynamics, contact changes, external collisions, payload shifts, or other events absent from policy training~\citep{haider2023out,nasvytis2024rethinking}. These failures are challenging because learned policies can continue to act confidently after entering unsupported regions of the state--action distribution.

Uncertainty-based methods such as MC Dropout, DropConnect, and ensembles estimate epistemic uncertainty through stochastic forward passes or model disagreement~\citep{mohammed2021benchmark}. While useful for detecting unfamiliar states, their inference cost scales with samples or models, and they may miss environment or transition shifts when the policy remains confidently wrong~\citep{prashant2025guaranteeing}. RAPT instead uses a single-pass probabilistic monitor: aleatoric uncertainty is represented through a heteroscedastic likelihood output, while epistemic-style deployment uncertainty is exposed by running an independent nominal-behavior model alongside the policy.

Generative, reconstruction-based, and time-series detectors learn nominal observations and flag low-likelihood inputs~\citep{park2018multimodal,sellam2025mamba,zhong2025patchad}. However, likelihood models can suffer from input-complexity bias~\citep{nalisnick2018deep}, while generic time-series methods often treat proprioceptive channels as independent signals rather than coupled robot states. Dynamics-based methods model transitions such as $P(s_{t+1}\!\mid s_t,a_t)$ and can detect mechanical or environmental perturbations~\citep{haider2023out,prashant2025guaranteeing}. Conformal prediction provides finite-sample coverage guarantees under exchangeability assumptions~\citep{xu2025can}, but high-frequency humanoid deployment requires thresholds across many dimensions, rare long-tailed nominal events dominate low-FPR behavior, and safety demands extremely low episode-level false-positive rates. RAPT therefore uses calibrated per-dimension and global predictive-deviation gates as a practical compromise for low-FPR real-time detection with localization.

Control-theoretic safety methods provide stronger guarantees when accurate dynamics and constraints are available~\citep{brunke2022safe,berkenkamp2017safe,chow2019lyapunov,cheng2019end,alshiekh2018safe}. However, learned humanoid controllers often operate in high-dimensional proprioceptive spaces where explicit dynamics are incomplete, difficult to identify, or too conservative for agile behavior. RAPT targets the intermediate regime where a controller is already trained, but residual sim-to-real mismatch and unforeseen real-world OOD events must still be detected, localized, and diagnosed. It complements methods that reduce the Sim-to-Real gap before deployment by monitoring the residual mismatch and deployment events that remain after transfer. Unlike pipelines trained mainly from small nominal real-world datasets, RAPT uses large-scale simulation to collect diverse nominal behavior for self-supervised detector training, while controlled OOD perturbations with known categories are used for evaluation and ablation in simulation. These labels are not required at deployment, but allow systematic testing of detection and diagnosis assumptions before adapting RAPT through brief nominal real-world calibration.

\section{Methodology}

\begin{figure*}[ht]
    \centering
    \includegraphics[width=0.9\textwidth]{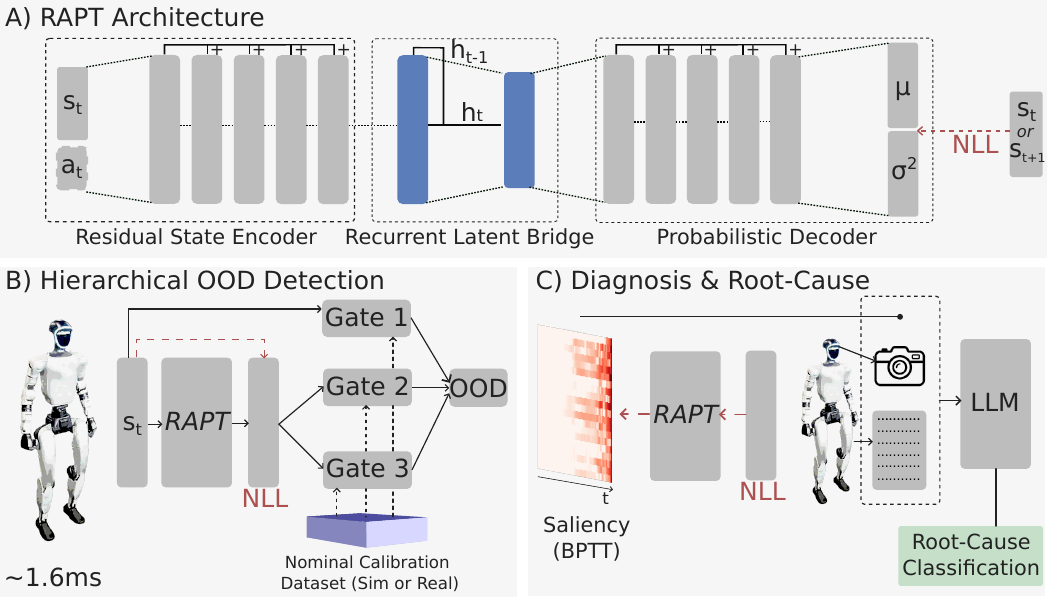}
    \caption{\textbf{RAPT Method Overview:} (A) RAPT OOD-detection architecture. (B) Hierarchical OOD pipeline using three statistical gates for real-time ($\sim$1.6ms) monitoring. (C) Detected anomalies trigger gradient-based saliency generation ($s_t$) for zero-shot diagnosis via a multi-modal LLM.}
    \label{fig:rapt_overview}
    \vspace{-1em}
\end{figure*}

RAPT is a lightweight monitor that runs alongside a pre-trained humanoid policy. The pipeline has four stages: (i) collect nominal execution data from expert policies in large-scale simulation; (ii) train a self-supervised probabilistic trajectory model; (iii) calibrate predictive-deviation thresholds using nominal simulation or verified real-world data; and (iv) diagnose detected failures using temporal saliency and semantic reasoning. Fig.~\ref{fig:rapt_overview} summarizes the pipeline; implementation details and hyperparameters are provided in the Supplementary Material.

\subsection{Nominal Data Collection}

For each task, we execute a pre-trained expert policy in NVIDIA Isaac Lab and collect nominal proprioceptive trajectories $\mathbf{o}_t \in \mathbb{R}^{d_{\text{obs}}}$ from the Unitree G1 humanoid. These simulation trajectories define the nominal behavior manifold that RAPT should preserve during deployment. All observations are normalized to zero mean and unit variance prior to training. Because the detector is trained independently of the policy, it can be attached to existing learned controllers without modifying the control architecture or reward function.

\subsection{Probabilistic Recurrent Trajectory Model}

RAPT is trained as a self-supervised recurrent reconstruction model over nominal execution. The architecture consists of a residual state encoder, a GRU-based temporal bridge, a latent bottleneck, and a probabilistic decoder. Given an observation history $\mathbf{o}_{\leq t}$, the encoder, GRU, and bottleneck produce a latent temporal representation $\mathbf{z}_t$, which the decoder maps to a diagonal Gaussian reconstruction distribution:
\begin{equation}
    [\boldsymbol{\mu}_t, \log \boldsymbol{\sigma}^2_t] = d_\theta(\mathbf{z}_t),
    \qquad
    \mathbf{z}_t = f_\phi(\mathbf{o}_{\leq t}).
\end{equation}
The model is trained by minimizing the negative log-likelihood (NLL) of nominal observations under this predicted distribution. This probabilistic formulation captures heteroscedastic aleatoric uncertainty: dimensions with naturally higher variability can be assigned larger predictive variance, while consistent dimensions are penalized more strongly when they deviate. We use denoising and a moderate latent bottleneck to discourage trivial identity reconstruction.

For velocity tracking, we also evaluate a forward-dynamics variant that conditions on the previous observation and action $(\mathbf{o}_{t-1}, \mathbf{a}_{t-1})$ and predicts $\mathbf{o}_{t}$ rather than reconstructing $\mathbf{o}_t$. This was the only task-dependent design choice in the ablation: forward dynamics performed best for velocity tracking, while pure reconstruction was preferred for the other tasks; the remaining components were consistently beneficial across tasks. Full architectural details, loss expressions, training hyperparameters, and ablations are provided in the Supplementary Material.

\subsection{Simulation-Trained, Deployment-Calibrated OOD Detection}

At deployment, RAPT evaluates the per-dimension predictive deviation of each incoming observation under the nominal trajectory model. Let $\ell_{t,i}$ denote the NLL contribution of observation dimension $i$ at time $t$, and let $\bar{\ell}_t$ denote the mean NLL across dimensions. Detection thresholds are calibrated using nominal data from either simulation or real-world operation, depending on the use case. When calibrated on a large nominal simulation batch, RAPT measures deviations from the simulation-trained behavior distribution, making residual Sim-to-Real mismatch visible during deployment. When calibrated on a brief verified nominal real-world run, RAPT adapts to static deployment-domain offsets caused by sensor noise, actuation latency, contact differences, and hardware-specific dynamics, enabling detection of new OOD events relative to nominal hardware behavior.

RAPT applies two gates for detection: a local gate that detects unusually large per-dimension deviations, and a global gate that detects broad shifts in the mean predictive deviation. In our implementation, thresholds are set using a max-plus-$k\sigma$ rule over the selected calibration data:
\begin{equation}
\label{eq:rapt_gate}
\text{Gate}_{1,2}(\mathbf{o}_t)=
\left(\exists i:\ell_{t,i}>\tau_i^{\max}+5\sigma_i\right)
\lor
\left(\bar{\ell}_t>\tau_{\text{global}}+3\sigma_{\text{global}}\right).
\end{equation}
Rare nominal spikes observed during calibration are treated as nominal, while persistent, localized, or distributed deviations beyond this envelope are flagged as OOD; local gates support channel-level localization and the global gate captures broader mismatch.

\subsection{Hybrid Safety Envelope}
RAPT is optionally paired with a deterministic range detector to provide a simple safety envelope for observations outside calibrated nominal bounds. The final anomaly decision is the logical OR of the local RAPT gate, global RAPT gate, and range detector; range-bound construction is provided in the Supplementary Material.

\subsection{Temporal Saliency for Failure Localization}

After an anomaly is detected, RAPT computes post-hoc temporal saliency to identify which observations contributed most strongly to the predictive deviation. Because the detector is recurrent, failures may accumulate over time before crossing the detection threshold. We therefore compute gradients of the final NLL with respect to a history window $\mathbf{X}_t=[\mathbf{o}_{t-H},\dots,\mathbf{o}_t]$, using integrated gradients through the recurrent model. This produces a spatio-temporal attribution map over observation dimensions and past timesteps, highlighting both \emph{where} and \emph{when} the execution first departed from nominal behavior.

Using the NLL as the saliency objective naturally weights attributions by the model's predictive uncertainty: deviations in high-confidence dimensions receive stronger attribution than deviations in noisy or intrinsically variable dimensions. This allows operators to distinguish sudden external disturbances, gradual actuator degradation, sensor drift, and other temporally structured failure modes. Full details of the saliency baseline, integration procedure, and visualization protocol are provided in the Supplementary Material.

\subsection{Semantic Failure Diagnosis}

Saliency localizes anomalous features but does not by itself provide a semantic explanation. To support post-hoc diagnosis, we provide a multi-modal LLM with structured diagnostic evidence, including temporal saliency summaries, joint-kinematic trajectories, task context, and, when available, a synchronized visual keyframe. The LLM performs zero-shot classification over a predefined set of failure categories, producing a likely failure hypothesis for operator review.

This diagnostic stage is not used for real-time safety triggering and is not intended to replace human judgment. Instead, it converts calibrated anomaly evidence into interpretable semantic labels that can accelerate post-incident debugging, such as distinguishing terrain/contact issues, external collisions, payload mismatch, sensor corruption, latency, or actuator faults. The failure taxonomy and prompt format are given in the Supplementary Material.

\section{Experimental Setup}
\label{sec:experimental_setup}

\subsection{Tasks, Policies, and Simulation Protocol}
\label{subsec:sim_protocol}
We evaluate RAPT on four Unitree G1 humanoid tasks: velocity-tracking locomotion, motion mimicry (Dance102 and Gangnam), and ballistic throwing. Locomotion and mimicry follow the \texttt{unitree\_rl\_lab} benchmark~\citep{unitree_rl_lab}, while throwing follows~\citep{munn2025whole}. For each task, we train five expert policies with different random seeds using GCR-PPO~\citep{munn2025scalable}.

Large-scale OOD evaluation is performed in NVIDIA Isaac Lab using $N_{\text{env}}=4096$ parallel environments per task. Each evaluation run uses a 50/50 split between nominal rollouts for false-positive evaluation and controlled OOD perturbations for true-positive evaluation. We inject 15 OOD categories spanning sensor faults, physics perturbations, actuator changes, initialization errors, and software/communication faults. A brief nominal calibration episode precedes evaluation to establish detector thresholds. Additional task, perturbation, and implementation details are provided in the Supplementary Material.

\subsection{Baselines}
We compare RAPT against representative high-frequency-compatible anomaly detectors: PatchAD~\citep{zhong2025patchad}, LSTM-VAE~\citep{park2018multimodal}, Deep SVDD~\citep{ruff2018deep}, and Isolation Forest~\citep{liu2008isolation}. All methods are trained on identical nominal data and evaluated on the same logs. Where source code was unavailable, we re-implemented methods following the original specifications and selected hyperparameters for stable 50\,Hz deployment under comparable computational budgets; full details are provided in the Supplementary Material.

\subsection{Real-World Deployment}
\label{subsec:real-world_deploy}

We deploy RAPT on a physical Unitree G1 humanoid at 50\,Hz. The real-world evaluation contains $78$ runs: $15$ nominal and $63$ anomalous. A brief calibration set of $3{\times}1$\,min nominal runs is used to set thresholds; nominal evaluation includes 1, 2, 4, and 10\,min runs, with half of the longer runs collected the day after calibration to test temporal robustness. Anomalous trials span contact, payload, actuation, latency, initialization, observation-ordering, pushes, collisions, motor failures, dynamics mismatch, and sensor-noise perturbations. All proprioceptive data and RAPT outputs are logged online, and baselines are evaluated offline on the same trajectories.

\subsection{Evaluation Metrics}
In simulation, we report AUROC and Safety Score, defined as True Positive Rate (TPR) at a fixed episode-level False Positive Rate (FPR) of $0.5\%$, reflecting the low false-positive requirements of real deployment. We also report per-step inference runtime to assess suitability for 50\,Hz control. Since all detectors operate at the same control frequency, detection delay is determined by anomaly-score dynamics rather than raw inference speed; category-specific delay statistics are reported in the Supplementary Material.

On hardware, we report trial-level confusion counts where available, together with TPR and FPR. For semantic failure diagnosis, we evaluate zero-shot classification accuracy on the subset of detected OOD trials with complete diagnostic labels and synchronized evidence.

\section{Results}

We evaluate RAPT along three axes: simulation detection accuracy, real-world generalization, and semantic failure diagnosis.
\subsection{Simulation Benchmarks}
\label{subsec:simulation_benchmarks}
We evaluate anomaly detection performance in simulation following the protocol described in Section~\ref{subsec:sim_protocol}. Results are summarized using inference latency, AUROC, and Safety Score (TPR @ 0.5\% FPR), as defined in the Experimental Setup.

\textbf{Performance Analysis:}
As shown in Table~\ref{tab:main_results}, RAPT attains the highest Safety Score and AUROC across all evaluated tasks while maintaining low inference latency (1.63\,ms, $<10\%$ of the 20\,ms control budget). Compared to the strongest baseline (LSTM-VAE), RAPT achieves a +0.34 absolute improvement in Safety Score at the fixed false-positive operating point.

Notably, the relative performance gap between RAPT and prior methods is more pronounced for Safety Score than AUROC. This suggests that several baselines achieve reasonable threshold-independent separability, but struggle with sensitivity when false positives are tightly constrained. We attribute this to overlap between in-distribution and out-of-distribution tails, which degrades precision at low false-positive rates.

\textbf{Hybrid vs. Model-Only Detection:}
To ensure fair comparison, all methods were evaluated both with and without the auxiliary Range Detector. As shown in the \emph{Model Only} and \emph{Hybrid} columns of Table~\ref{tab:main_results}, shallow baselines (e.g., Isolation Forest, Deep SVDD) rely heavily on explicit range checks, exhibiting reduced performance when evaluated in isolation. In contrast, RAPT remains robust without the range detector (0.75 vs.\ 0.72 Safety Score), indicating that it internalizes much of the valid state manifold structure. The small drop in the hybrid setting is attributable to additional false positives introduced by conservative range thresholds.

\begin{table*}[t]
\centering
\caption{\textbf{Safety \& Performance Comparison.} Safety Score (TPR @ 0.5\% FPR) reported per task, with aggregate AUROC. RAPT attains the highest values on both metrics across all tasks. Model Only and Hybrid denote performance without and with the Range Detector, respectively. Results averaged over 5 seeds (± std). * indicates RAPT trained with a forward dynamics objective; all others use reconstruction.}
\vspace{0.25em}
\label{tab:main_results}
\resizebox{\textwidth}{!}{%
\begin{tabular}{l | c c | c c c c | c c}
\toprule
& \multicolumn{2}{c|}{\textbf{Global Metrics}} & \multicolumn{4}{c|}{\textbf{Safety Score (TPR @ 0.5\% FPR) by Task} $\uparrow$} & \multicolumn{2}{c}{\textbf{Ablation}} \\
\textbf{Method} & \textbf{Latency} & \textbf{Avg AUROC} & \textbf{Throwing} & \textbf{Velocity} & \textbf{Mimic (Dance)} & \textbf{Mimic (Gangnam)} & \textbf{Model Only} & \textbf{Hybrid} \\
\midrule
Isolation Forest & 4.32 ms & 0.69 & 0.24 \scriptsize{±0.04} & 0.34 \scriptsize{±0.03} & 0.42 \scriptsize{±0.01} & 0.38 \scriptsize{±0.00} & 0.18 & 0.34 \\
PatchAD & 11.45 ms & 0.73 & 0.14 \scriptsize{±0.01} & 0.16 \scriptsize{±0.03} & 0.17 \scriptsize{±0.04} & 0.16 \scriptsize{±0.03} & 0.18 & 0.16 \\
Deep SVDD & 0.45 ms & 0.67 & 0.29 \scriptsize{±0.03} & 0.31 \scriptsize{±0.10} & 0.41 \scriptsize{±0.01} & 0.37 \scriptsize{±0.01} & 0.14 & 0.34 \\
LSTM-VAE & 1.77 ms & 0.77 & 0.30 \scriptsize{±0.02} & 0.36 \scriptsize{±0.02} & 0.44 \scriptsize{±0.01} & 0.42 \scriptsize{±0.01} & 0.32 & 0.38 \\
Ours (RAPT) & 1.63 ms & \textbf{0.92} & \textbf{0.72} \scriptsize{±0.05} & 
\textbf{0.74}$^*$ \scriptsize{±0.02} & \textbf{0.67} \scriptsize{±0.08} & \textbf{0.75} \scriptsize{±0.02} & \textbf{0.75} & \textbf{0.72} \\
\bottomrule
\end{tabular}%
}
\vspace{-1em}
\end{table*}

\subsection{Ablation Studies}
\label{subsec:ablation}

We analyze key design choices using a Taguchi L12 orthogonal array over 48 models, with full results in the Supplementary Material. The dominant factor is inference logic: per-dimension maximum aggregation improves Safety Score by +14.3\% over global mean aggregation, indicating that OOD deviations are often localized to a subset of channels. Probabilistic NLL scoring improves robustness under heteroscedastic noise (+10.3\%), while temporal recurrence (+8.5\%), low latent compression (+7.8\%), and residual connections (+4.5\%) provide additional gains. These results motivate the final RAPT design. The results challenge the assumption that a single global anomaly score is sufficient for robot deployment: under strict low-FPR constraints, failures in high-DOF humanoid control are often expressed as localized deviations in a small subset of channels, making calibrated per-dimension evidence critical for both detection and interpretation.





\subsection{Real-World Deployment}
\label{subsec:real_world_deployment}

We evaluate RAPT on physical hardware using the protocol in Section~\ref{subsec:real-world_deploy}. As shown in Table~\ref{tab:real_world_results}, RAPT+Range achieves the strongest real-world performance, detecting 56 out of 63 anomalous runs with one false positive over 15 nominal trials (88.9\% TPR, 6.7\% episode-level FPR); the single false positive occurred during a deliberately challenging 10\,min nominal run. PatchAD is omitted from the expanded table because many safety-terminated runs were shorter than its required context window; on the 27-run subset where PatchAD was applicable, it achieved 41.6\% recall, compared with 66.7\% for RAPT and 75.0\% for RAPT+Range, all without false positives. RAPT detects dynamic interaction faults that do not necessarily violate explicit bounds, while the range detector captures hard state-space violations; their combination improves robustness across physical perturbations. Beyond quantitative detection, RAPT also flagged silent configuration errors during integration, such as under-tuned wrist PD gains, highlighting its use as a deployment verification tool.

\subsection{Automated Root-Cause Diagnosis}
\label{subsec:automated_rca}

We evaluate the diagnostic pipeline on 16 real-world failure logs with complete diagnostic data following successful anomaly detection by RAPT (Hybrid). The subset is intended to test whether temporal saliency through the recurrent detector provides useful diagnostic evidence, so we focus on failures that evolve over time rather than immediate, protocol-defined violations whose category is typically evident from the test condition. These cases include external disturbances, payload mismatch, terrain/contact changes, and physical obstructions, where diagnosis requires interpreting temporal proprioceptive and kinematic evidence. Each failure is classified into one of 21 predefined categories (see Supplementary Material), and results are reported for Proprioceptive Only and Multi-Modal variants, the latter using a single time-synchronized visual key-frame.

As shown in Table~\ref{tab:rca_results}, the Multi-Modal configuration improves both Top-1 and Top-3 accuracy relative to the proprioceptive-only baseline. Visual context helps disambiguate external disturbances from internal system faults, such as obstacle contact versus motor-dynamics mismatch.

From an operational perspective, narrowing a failure to the relevant subsystem or signal group can substantially reduce post-incident debugging time. The diagnostic pipeline helps distinguish localized faults, such as an individual motor or sensor channel, from broader semantic shifts caused by external contact, payload mismatch, latency, or controller/configuration errors. In practice, this helped identify silent sim-to-real discrepancies, such as mismatched PD gains or implementation-level inconsistencies in the deployed state pipeline.

\begin{table*}[t!]
\centering
\caption{\textbf{Real-World Anomaly Detection on Unitree G1.} Results over $N=78$ trials: $15$ nominal and $63$ anomalous. PatchAD is omitted because many safety-terminated anomalous runs were shorter than its required context window.}
\vspace{0.15em}
\label{tab:real_world_results}
\scriptsize
\setlength{\tabcolsep}{5pt}
\renewcommand{\arraystretch}{0.82}
\begin{tabularx}{\textwidth}{X | c c | c c c c}
\toprule
\textbf{Method} & \textbf{TPR} $\uparrow$ & \textbf{FPR} $\downarrow$ & \textbf{TP} & \textbf{FN} & \textbf{TN} & \textbf{FP} \\
\midrule
SVDD               & 28.3\% & 60.0\% & 18 & 45 & 6  & 9 \\
Isolation Forest   & 26.0\% & 13.3\% & 16 & 47 & 13 & 2 \\
LSTM-VAE           & 28.6\% & 13.3\% & 18 & 45 & 13 & 2 \\
\midrule
RAPT (Ours)        & 85.7\% & 6.7\%  & 54 & 9  & 14 & 1 \\
\textbf{RAPT + Range (Hybrid)} & \textbf{88.9\%} & \textbf{6.7\%} & \textbf{56} & \textbf{7} & \textbf{14} & \textbf{1} \\
\bottomrule
\end{tabularx}
\vspace{-0em}
\end{table*}


\begin{table*}[t!]
\vspace{-1em}
\centering
\caption{\textbf{Semantic Failure Diagnosis Accuracy.} Top-1 and Top-3 classification accuracy across 16 real-world failure scenarios.}
\vspace{0.15em}
\label{tab:rca_results}
\scriptsize
\setlength{\tabcolsep}{4pt}
\renewcommand{\arraystretch}{0.82}
\begin{tabularx}{\textwidth}{X | c c | c c}
\toprule
\textbf{Method} & \multicolumn{2}{c|}{\textbf{Top-1 Accuracy}} & \multicolumn{2}{c}{\textbf{Top-3 Accuracy}} \\
\cmidrule(lr){2-3} \cmidrule(l){4-5}
& \textbf{Count} & \textbf{\%} & \textbf{Count} & \textbf{\%} \\
\midrule
Proprioceptive Only (RAPT) & 12 / 16 & 75.0 & 14 / 16 & 87.5 \\
\textbf{RAPT + Visual Keyframe} & \textbf{14 / 16} & \textbf{87.5} & \textbf{16 / 16} & \textbf{100.0} \\
\midrule
\textit{Improvement} & 2 / 16 & 12.5 & 2 / 16 & 12.5 \\
\bottomrule
\end{tabularx}
\vspace{-2em}
\end{table*}
\vspace{-0.5em}
\section{Limitations}
RAPT has several limitations. Hardware evaluation remains constrained by safety, time, and platform availability: trials cover diverse OOD events, but only on a single humanoid over short-term deployments. RAPT is also a monitor rather than a formal safety guarantee: it detects deviations from calibrated nominal behavior and triggers predefined responses, but does not synthesize safe controls or prove constraint satisfaction. Calibration can trade false positives against sensitivity, with more conservative thresholds reducing false alarms at the cost of delayed or missed detections. Finally, semantic diagnosis is post-hoc and assistive, so ambiguous or compound failures may still require expert interpretation.

\section{Discussion and Conclusion}
We presented RAPT, a lightweight self-supervised monitor for detecting and diagnosing OOD behavior during deployment of learned humanoid policies. RAPT learns a probabilistic spatio-temporal model of nominal execution and uses calibrated predictive deviations for high-frequency, low-FPR detection. Unlike binary anomaly detectors, RAPT provides per-dimension evidence that localizes when and where execution departs from nominal behavior, enabling temporal saliency and semantic diagnosis.

Across simulation and real-world experiments, RAPT outperforms high-frequency-compatible baselines while remaining real-time deployable. The results suggest that RAPT is useful for both safety triggering and deployment debugging, complementing domain randomization and system identification by monitoring residual mismatch and unforeseen OOD events after transfer.

\clearpage
\acknowledgments{If a paper is accepted, the final camera-ready version will (and probably should) include acknowledgments. All acknowledgments go at the end of the paper, including thanks to reviewers who gave useful comments, to colleagues who contributed to the ideas, and to funding agencies and corporate sponsors that provided financial support.}


\bibliography{example}  

\end{document}